\documentclass[11pt]{article}
\usepackage[hyperref]{ccl2024-en}
\usepackage{times}
\usepackage{url}
\usepackage{latexsym}
\usepackage{fancyhdr}

\usepackage{natbib}
\usepackage{bigstrut}
\usepackage{pifont}
\usepackage{amsmath} 
\usepackage{array}
\usepackage{tablefootnote}
\usepackage{booktabs}
\usepackage{multirow}
\usepackage{tabularray}
\usepackage{tabularx}
\usepackage{graphicx}
\usepackage{adjustbox}
\usepackage{CJKutf8}
\usepackage{tcolorbox}
\UseTblrLibrary{booktabs, counter}
\usepackage[normalem]{ulem}
\usepackage{soul}
\usepackage{arydshln}
\usepackage[inkscapelatex=false]{svg}
\usepackage{pdfpages}

\newcommand{\tick}{\ding{51}\quad}
\newcommand{\fork}{\ding{55}\quad}

\newcolumntype{C}[1]{>{\centering\arraybackslash}m{#1}}
\usepackage{soul}

\newtcolorbox{violate}{colback=red!10!white,colframe=red!50!black}
\newtcolorbox{comply}{colback=green!10!white,colframe=green!50!black}

\usepackage{color}

\title{Do Large Language Models Understand Conversational Implicature --\\
A case study with a Chinese sitcom
}

\author{Shisen Yue \quad Siyuan Song \quad Xinyuan Cheng \quad Hai Hu* \\
        School of Foreign Languages, Shanghai Jiao Tong University \\ \texttt{shisenyue@gmail.com}, \texttt{\{sjtusongsy2022, 0106cxy, hu.hai\}@sjtu.edu.cn}\\
        *corresponding author}

\CJKnospace

\date{}
\begin{document}
\blfootnote{\textcopyright 2024 China National Conference on Computational Linguistics \\ Published under Creative Common Attribution 4.0 International License}\maketitle

\begin{abstract}
Understanding the non-literal meaning of an utterance is critical for large language models (LLMs) to become human-like social communicators. In this work, we introduce \mbox{SwordsmanImp}, the first Chinese multi-turn-dialogue-based dataset aimed at conversational implicature, sourced from dialogues in the Chinese sitcom \textit{My Own Swordsman}. It includes 200 carefully handcrafted questions,  all annotated on which Gricean maxims have been violated.  We test eight close-source and open-source LLMs under two tasks: a multiple-choice question task  and an implicature explanation task. Our results show that GPT-4 attains human-level accuracy (94\%) on multiple-choice questions. CausalLM demonstrates a 78.5\% accuracy following GPT-4. Other models, including GPT3.5 and several open-source models, demonstrate a lower accuracy ranging from 20\% to 60\% on multiple-choice questions. Human raters were asked to rate the explanation of the implicatures generated by LLMs on their reasonability, logic and fluency. While all models generate largely fluent and self-consistent text, their explanations score low on reasonability except for GPT-4, suggesting that most LLMs cannot produce satisfactory explanations of the implicatures in the conversation. Moreover, we find LLMs' performance does not vary significantly by Gricean maxims, suggesting that LLMs do not seem to process implicatures derived from different maxims differently. Our data and code are available at \href{https://github.com/sjtu-compling/llm-pragmatics}{https://github.com/sjtu-compling/llm-pragmatics}.
\end{abstract}

\section{Introduction}

The complexity of communication is largely epitomized by indirect, or non-literal utterances. A common instance is hinting at a busy schedule as a polite refusal to engage in an unwanted activity. 
How such implied meaning is understood in human communication has lone been a key subject of investigation in pragmatics research~\citep{grice1975logic, searle1980speech, brown1987politeness, wilson2006relevance}. 

\begin{figure*}[!th]
    \centering
    \includegraphics[width=\textwidth]{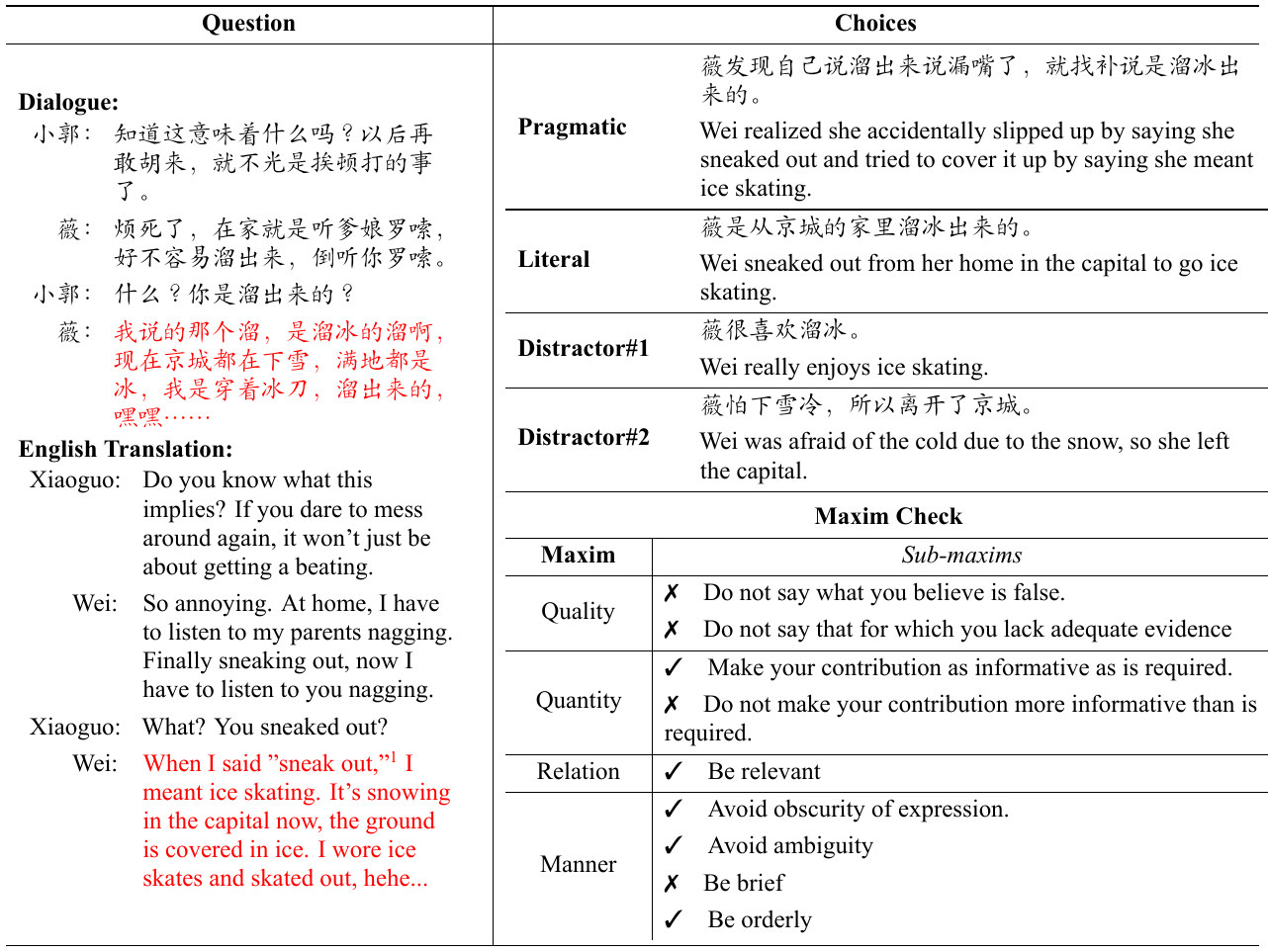}
    \caption{An example entry in our dataset. The tick (\ding{51}) and fork (\ding{55}) denotes if the target sentence, marked in red, comply with or violate the sub-maxim. This entry belongs to the classes of Quality, Quantity and Manner. ``sneak out" and ``skate" translate to the same Chinese character \begin{CJK*}{UTF8}{gkai}``溜''\end{CJK*}.}
    \label{tab:eg:dialogue:intro}
\end{figure*}

Evaluating the pragmatic understanding ability of large language models (LLMs) has drawn considerable attention in recent years as LLMs show remarkable ability for language understanding. 
Recent studies have evaluated LLMs' pragmatic reasoning in multiple aspects, including scalar inference \citep{hu2023expectations}, discourse connectives \citep{pandia2021pragmatic}, gradable adjectives \citep{lipkin2023evaluating} and conversational implicatures \citep{qiu2023does, kim2023pope, ruis2022large, finegrainedPragmatic, zheng2021grice}. 

However, the above-mentioned evaluation are primarily in English, leaving a gap for pragmatic understanding in other languages. Moreover, previous conversational reasoning datasets involve few turns of dialogue, while 
our daily communication usually involves much more context and turn taking. 
In addition, previous studies mostly employ binary- or multiple-choice questions~\citep{finegrainedPragmatic,ruis2022large}, which is inadequate in the era of LLM where it is possible to ask LLMs to generate explanatory text of the situation and directly evaluate its quality. 

To bridge the aforementioned gaps, we present SwordsmanImp, the first Chinese multi-turn-dialogue-based dataset aimed at conversational implicature. It consists of 200 dialogues carefully excerpted by linguistic experts from the Chinese sitcom \textit{My Own Swordsman} \begin{CJK*}{UTF8}{gkai}(武林外传)\end{CJK*}. Figure~\ref{tab:eg:dialogue:intro} shows an example dialogue in the dataset. Each dialogue contains one sentence that carries a non-literal meaning. We provide four well-organized interpretations of this sentence, including a pragmatic meaning, a literal understanding, and two incorrect inferences that involve related information (distractors). 

From a linguistic perspective, many pragmatic inferences, especially conversational implicature, arise because the speaker has violated one or more maxims in the cooperative principle~\citep{grice1989retrospective}. For example: when asked ``Is John in the office?'', Sam replied ``It's Saturday, you know''. This violates the Maxim of Relation since the reply is not directly related to the question, which then gives rise to an implicature: ``John never works on weekends, so he is not in the office''. 
Meanwhile, it has been shown that the difficulty in detecting conversational implicature hinges crucially on which maxims have been violated~\citep{engelhardt2006speakers, rubio2019overinformative, okanda2015understanding, panzeri2021children}. Thus we meticulously annotated each conversation to determine which of the four maxims in the cooperative principle have been violated. 

In this work, we ask the following research questions:

(1) How do state-of-the-art LLMs perform in interpreting implicatures written in Chinese, in multiple choice style? (Section~\ref{sec:exp1})

(2) What are the error patterns of LLMs on multiple choice questions? (Section~\ref{sec:exp1})

(3) How do human participants rate the interpretation of conversational implicature generated by LLMs? (Section~\ref{sec:exp2})

We first review relevant research in Section~\ref{sec:related} and then describe how we built our dataset SwordsmanImp in Section~\ref{sec:corpus}. In Section~\ref{sec:exp1}, we measure the models' accuracy on multiple-choice questions derived from our dataset. We test four models from GPT series \citep{radford2018improving}, four open-source models. In Section~\ref{sec:exp2}, we use five models from Experiment 1 to generate pragmatic interpretations of 32 dialogues, and then we invite human participants to rate these interpretations in three dimensions: reasonability, logic and fluency.

We find that GPT-4 attains a human-level accuracy in multiple-choice questions. CausalLM (14B) also reaches a high accuracy as 78.5\% on this task. Moreover, our results show that LLMs' performance does not vary with respect to different conversational maxims. Furthermore, we reveal that while in general fluent and self-consistent, occasional confused logic, redundant information and unexpected language tokens are the main reasons for their explanations to be underrated by human evaluators.

\section{Related Work}\label{sec:related}

\begin{table*}[t]
\small
\centering
    \begin{adjustbox}{max width=\textwidth}
        \begin{tblr}{Q[l, m]Q[l, m]Q[l, m]Q[c, m]Q[l, m]}
        \toprule
            Dataset         &        Task        &     Context       &         Manually        &   Language \\
        \midrule
            CoQa \citep{reddy2019coqa} & conversational QA & Paragraph & \tick & English \\
            Narrative-QA \citep{kocisky-etal-2018-narrativeqa} & Reading comprehension & Documents & \tick & English \\
            DREAM \citep{sun-etal-2019-dream} & QA & Free-form, multi-turn dialogue & \tick & English \\
            MuTual \citep{cui-etal-2020-mutual} & Next utterance prediction & Free-form, multi-turn dialogue & \tick & English \\
            GRICE \citep{zheng2021grice} & Gricean implicature recovery, QA & Fixed form, multi-turn dialogue & \tick & English \\
            PragMega \citep{floyd2023pragmega} & ToM QA & Paragraph & \tick & English \\
            DiPlomat \citep{li2023diplomat} & Pragmatic identification and reasoning, QA & Free-form, multi-turn dialogue & \fork & English \\
            LUDWIG \citep{ruis2022large} & Implicature QA & Single-turn dialogue & \tick & English \\
             Douban \citep{wu-etal-2017-sequential} & Next utterance prediction & Free-form, multi-turn dialogue & \fork & Chinese \\
        \midrule
            SwordsmanImp (This work) & Gricean Implicature QA & Free-form, multi-turn dialogue & \tick & Chinese \\
        \bottomrule
        \end{tblr}
    \end{adjustbox}
\label{tab:comparison:datasets}
\caption{Comparing our datasets and the existing datasets. ``Manually'' indicates whether the questions or answers are written partly or entirely by human.}
\end{table*}

Understanding non-literal meanings has long been considered a difficult task for language models. Previous studies have explored the capabilities of language models in recognizing metaphors \citep{wachowiak2023does,neidlein2020analysis}, humor \citep{hessel2022androids,jentzsch2023chatgpt,chen2023can} and social commonsense \citep{sap-etal-2019-social}. Broad-scale datasets aimed at pragmatic reasoning collect data mainly through crowdsourcing or crawling from the internet \citep{reddy2019coqa, kocisky-etal-2018-narrativeqa}. Some others transform the existing tests for humans to train and evaluate language models \citep{cui-etal-2020-mutual, sun-etal-2019-dream}. 
While the above-mentioned datasets have comprehensive coverage for pragmatic understanding, they are not ideal resources for evaluation of a specific type of pragmatic knowledge, that is, conversational implicatures derived by violation of the Gricean maxims.
In our work, we craft our dataset manually to make sure that each implicature can be classified to one or more Gricean maxims. This enables us to perform a fine-grained analysis of LLMs' understanding of this particular pragmatic inference.

Previous attempts in evaluating LLMs' pragmatic understanding guided by the Gricean maxims or cooperative principles do not have multi-turn dialogues from real-world situations. 
Formalized under the theory of cooperative principles \citep{grice1975logic}, \citet{zheng2021grice} automatically generated their dataset GRICE through a set of well-defined grammar rules. 
This grammar-based method ensures a good control for the pragmatic cues to be aligned with the Gricean maxims. However, dialogues with fixed syntactic structures hardly represent the complexity of conversation in daily communication. 
In this study, we fill the gap by providing the multi-turn dialogue directly to the model.

Last but not least, there is a scarcity in Chinese evaluation resources for pragmatic inferences. Previous large-scale conversational datasets in Chinese use text sources from Sina Weibo \citep{shang-etal-2015-neural}, Douban conversational corpus \citep{wu-etal-2017-sequential} and E-commerce Dialogue corpus \citep{zhang-etal-2018-modeling}. However, all of them are not specifically aimed to test LLMs' pragmatic understanding. 
To address this gap, we select the Chinese sitcom \textit{My Own Swordsman} as the source to formulate pragmatic questions and construct a Chinese dataset for evaluating LLMs' pragmatic reasoning ability.

\section{Dataset Construction}\label{sec:corpus}

\subsection{Data source} 
It is not easy to find naturally occurring, high-quality, multi-turn dialogues. Following previous literature that uses situational comedies (sitcom) as sources \citep{wang2017binge,wu2021mumor,patro2021multimodal}, we chose the Chinese sitcom \textit{My Own Swordsman} \begin{CJK*}{UTF8}{gkai}(武林外传)\end{CJK*} as our data source, because of its abundance of conversational implicature, well-written dialogues and the uniquely Chinese background. 
Set in the Ming dynasty of China, this sitcom narrates the daily life of a group of people working in a Chinese motel. We believe it will be a unique resource for evaluating LLMs' pragmatic inference ability in both the Chinese language and the Chinese context.

\subsection{Annotation procedure}

\paragraph{Implicature identification and classification}

Three authors of this paper with at least two years of linguistics training went through the script of the sitcom to select conversations that contain conversational implicatures. 
Conversations with multiple turns were selected based on whether any conversational maxim is violated on purpose \citep{grice1975logic}. For each conversation, we performed the cancellation test\footnote{A test to diagnose the conversational implicature by encoding semantically the negation of the target meaning. If the result seems consistent, then the target meaning is likely an implicature.} to ensure that a conversational implicature rather than a semantic entailment is involved~\citep{hirschberg1985theory}. 
Multiple turns were included for each data entry to ensure that even if one has not seen the sitcom, one could still understand the conversation just from the snippet we selected. 

Then, the same three authors classified all chosen dialogues according to the conversational maxims they violate. The criteria of these maxims are drawn from  \citet{grice1975logic}. To allow for a more fine-grained classification, we employ sub-maxims as criteria, assessing if the target sentence fulfills each requirement individually. An utterance is considered to have violated a maxim if it infringes on any sub-maxim. Moreover, a dialogue might belong to multiple classes according to the sub-maxims the utterance violates. An illustrative dataset entry featuring a dialogue, four interpretations and a class is presented in 
Figure~\ref{tab:eg:dialogue:intro}.

\paragraph{Writing four interpretations}
Next, we construct the four interpretations of the sentence that carries implicature as the four choices: the pragmatic interpretation (the correct one), the literal interpretation, and two distractors with interpretations related to the context.\footnote{The distractors can be understood as ``neutral'' statements in the Natural Language Inference task~\citep{snli}.} The pragmatic meaning are provided based on human commonsense understanding. We construct the literal meaning by rephrasing the target sentence with concrete and unambiguous expressions. We generate incorrect inferences as distractors, which are relevant to the the topic of the conversation. The character names referred to in the four interpretations are aligned with the mentions in the dialogue clip. 

\paragraph{Verification}
We hired three PhD students in Linguistics to complete a multiple-choice question task, which is to choose the pragmatic understanding of the sentence from the shuffled four interpretations for all dialogues in our dataset. The students are required to complete the task independently without a time limit. After they finish, we invite them to discuss about their wrong answers and their reasoning process. This validation process guarantees that the provided pragmatic interpretation is closely aligned with the commonsense intuitive understanding and can be deduced from the limited context. Additional information that is necessary to pragmatic reasoning, such as relationships between characters, their personality, social background, and multi-modal information is supplemented in the bracket at the beginning of the dialogue. Besides, implicatures derived from vernacular, slang, and network-specific jokes have been filtered out from our dataset.

\subsection{Obtaining human score}
To compare the results with human performance, we ask 10 native speakers to work on 32 questions randomly sampled from the dataset and they achieve an average accuracy of 93.1\%. The questionnaire include the same number of each type of questions (i.e. Gricean maxims violated in the dialogue). The participants were all undergraduate students from a Chinese University, and they were compensated for their 
involvement in the experiment. Only 32 questions were used when estimating human score, because asking each annotator to solve 200 questions is implausible. The human accuracy should thus be regarded as an approximation of human performance.

\subsection{Resulting corpus: SwordsmanImp}

The final SwordsmanImp corpus contains 200 manually curated questions across four types categorized according to Gricean maxims. Each entry contains a multi-turn dialogue and four interpretations of the target sentence as choices (see Table~\ref{tab:meta-data}). 

\begin{table*}[t]
    \centering
    \small
\begin{adjustbox}{max width=\textwidth}
\begin{tabular}{llllll}
\toprule
&\textbf{Total}&\textbf{Quality}&\textbf{Quantity}&\textbf{Relevance}&\textbf{Manner}\\
\midrule
\# of questions&200&76&33&71&62\\
\# of turns per dialogue &6.80&7.84&5.91&6.23&6.35\\
Avg. dialogue length&158.22&184.53&143.67&147.20&152.79\\
Avg. Utterance length &23.27&23.53&24.31&23.64&24.04\\
Avg. Answer length &15.08&14.47&14.85&15.82&14.86\\

\bottomrule
\end{tabular}
\end{adjustbox}
\caption{Question numbers, average number of Chinese characters contained in each dialogue and utterance, and the number of turns per dialogue in our dataset.}
\label{tab:meta-data}
\end{table*}

\section{Experiment 1: multiple-choice question for LLMs} \label{sec:exp1}

In this experiment, the models will see the dialogue and the four interpretations we manually created. The task is to choose the correct interpretation (i.e., the pragmatic one) of the utterance that contains an implicature.   

\subsection{Models}
\label{sec:models-in-exp1}

We experiment with eight models in this experiment. 
The first four models are from the OpenAI GPT family, which are tested through the OpenAI API:
text-davinci-002, text-davinci-003, GPT-3.5-turbo and GPT4.\footnote{The four OpenAI models are evaluated on November 15th, 2023.
} 
We also examine four open-source models using the Transformers library \citep{wolf-etal-2020-transformers} from Huggingface\footnote{\url{https://huggingface.co}}. We evaluate Chinese-Alpaca-2-13B \citep{cui2023efficient},  OpenBuddy-Llama2-13B\footnote{\url{https://huggingface.co/OpenBuddy/openbuddy-llama2-13b-v8.1-fp16}} (based on Llama2 \citep{touvron2023llama}), CausalLM-13B (based on Llama2 and Qwen~\citep{qwen}), as well as BLOOMZ-7.1B from the BLOOM series~\citep{muennighoff2022crosslingual}.  

\begin{figure*}[t]  
 \centering  
 \includegraphics[width=\textwidth]{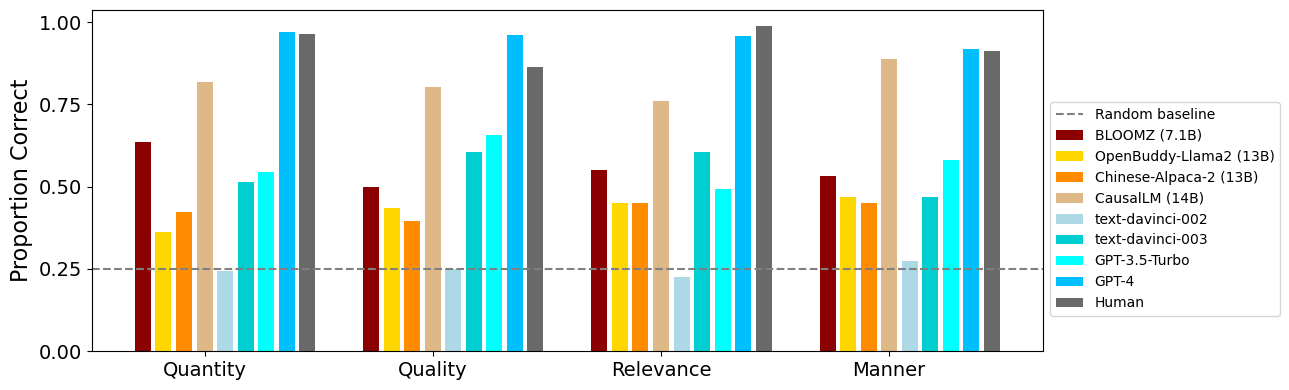}  
 \caption{Performance of models across question types in the multiple choice scenario. Dashed line represents chance accuracy}
 \label{fig:by_maxim}  
\end{figure*}

\begin{figure*}[t]  
\centering
 \begin{adjustbox}{max width=0.8\textwidth}
     \includegraphics[width=\textwidth]{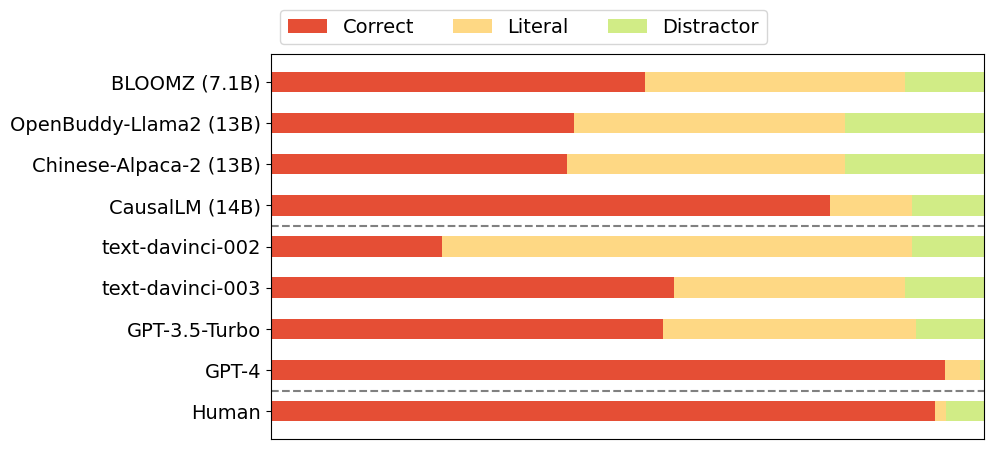}  
 \end{adjustbox}
 \caption{Frequency of each option in models' answers.}
 \label{fig:model-ans-freq}  
\end{figure*}

\subsection{Evaluation Protocol}
For both open-source and close-source models, we use zero-shot prompts to simulate the real-life situations where humans come across these non-literal expressions daily. 
However, we use two evaluation protocols depending on whether we have access to the probability/logits of all tokens in the vocabulary at a given output position.
For close-source models, we ask it to generate the answer and manually go over the generated text to determine which interpretation is chosen. 
For open-source models, we follow established practices in LLM evaluation~\citep{hu2023expectations,mmlu,li2023cmmlu} where we compute the logits of the four tokens (``A'', ``B'', ``C'', ``D'') after ``\begin{CJK*}{UTF8}{gkai}答案：\end{CJK*}''(answer:), and choose the one with the highest logit as model prediction (see an illustration in Appendix~\ref{app:prompt:exp1}.).

\begin{table}[t]
    \centering
\small
\begin{adjustbox}{max width=0.5\textwidth}
\begin{tabular}{lc}
\toprule
\textbf{Subject}&\textbf{Acc (\%)}\\
\midrule
Random baseline&25\\
\midrule
BLOOMZ (7.1B)&52.5\\
OpenBuddy-Llama2 (13B)&42.5\\
Chinese-Alpaca-2 (13B)&41.5\\
CausalLM (14B)&78.5\\
text-davinci-002&24.0\\
text-davinci-003&56.5\\
GPT-3.5-Turbo&55.0\\
GPT-4&94.0\\\midrule
Human&93.1\\
\bottomrule
\end{tabular}
\end{adjustbox}
\caption{Accuracy of language models and humans in experiment 1: multiple choice questions.}
\label{tab:total_acc}
\end{table}

\subsection{Results}
\label{sec:exp1:res}
The overall accuracy of all models and human participants is shown in Table~\ref{tab:total_acc}. 
We observe that GPT-4 achieves the highest accuracy of 94\%, showing a performance on par with human participants. CausalLM (14B) follows with 78.5\% accuracy. Other models exhibit difficulty in identifying the pragmatic meaning of the 
line containing an implicature, with accuracy ranging from 20\% to 60\%. Text-davinci-002 even fails to perform above chance (25\%). 
\textbf{This suggests that for most models we tested, there is still room for improvement in their understanding of implied meaning.
}\footnote{We also evaluate Baichuan2-13B-Chat and InternLM-Chat-20B (in half precision) with evaluation paradigm for close-source models. Their accuracy are separately 43\% and 62\%.} 

Table~\ref{fig:by_maxim} shows that models' performance grouped by the Gricean maxims violated in the dialogue. 
Overall, certain models demonstrate proficiency in answering questions related to a set of maxims, while others excel in different ones. \textbf{We do not observe a uniform pattern indicating a particular strength or weakness in any of these maxims across the models.} This lack of uniformity is 
also observed
in the results obtained from human participants who worked on the 32 sampled questions. 
Specifically, performance of text-davinci-002 is near to chance in all four conditions, demonstrating its incapacity in interpreting conversational implicature. 
The performance of OpenBuddy-Llama2 (13B) and Chinese-Alpaca-2 (13B) are comparable, both below 50\% accuracy, with the two models demonstrate predominance in dealing with different types of questions. The accuracy of GPT-3.5-Turbo, text-davinci-003 and BLOOMZ (7.1B) are in the same range, and they also demonstrate different performance orders in different conditions. It's noteworthy that BLOOMZ (7.1B) outperforms the two 13B models in all conditions. CausalLM (14B) demonstrates an accuracy close to human-level and the best performance within all open-source models tested in this experiment. GPT-4 attains the highest accuracy of 94.0\%, surpassing 90\% across all categories of questions. GPT-4's performance, although slightly higher than the human score, is not statistically different from the human accuracy ($p=0.802 > 0.05$ by a two-tailed t-test).

The distribution of the interpretations chosen by the models is shown in Figure~\ref{fig:model-ans-freq}, where red means that the model has chosen the correct answer, i.e., pragmatic interpretation, while yellow corresponds to the literal meaning, and green the two distractors. 
The performance of two 13B models shows a higher frequency of choosing distractors. 
This could possibly indicate a 
that these two models 
are easily sidetracked by irrelevant information in the context. We also observe that as the GPT models evolve, they have a higher chance of 
distinguishing literal meaning from implied meaning, culminating in the considerably low ratio of literal understanding in GPT-4. 

\section{Experiment 2: evaluating the quality of explanations generated by LLMs}\label{sec:exp2}

\begin{table*}[t]
\centering
\begin{adjustbox}{max width=\textwidth}
\begin{tabular}{lcccc}
\toprule
{}&\textbf{Reasonability}&\textbf{Logic}&\textbf{Fluency}&\textbf{Avg. response length}\\
\midrule
GPT-4&$4.24\pm0.68$&$4.65\pm0.39$&$4.91\pm0.13$&114.44\\
GPT-3.5-Turbo&$3.17\pm1.30$&$4.09\pm0.77$&$4.86\pm0.21$&125.41\\
Chinese-Alpaca-2 (13B)&$2.34\pm1.10$&$3.45\pm0.82$&$4.72\pm0.39$&156.19\\
CausalLM (14B)&$2.33\pm1.03$&$3.48\pm0.67$&$4.13\pm1.01$&147.41\\
Openbuddy-Llama2 (13B)&$2.11\pm0.99$&$3.55\pm0.71$&$4.52\pm0.65$&153.56\\
\bottomrule
\end{tabular}
\end{adjustbox}
\caption{Models' mean scores in three dimensions with standard deviation and the average number of Chinese characters in their responses.}
\label{tab:models-scores-in-3D}
\end{table*}

In the previous experiment, we ask LLMs to choose one answer from four choices.
In this experiment, we design open-ended questions where the models are asked to generate explanations of the implicature, which will then be evaluated manually by native speakers of Chinese, based on the reasonability/reasonableness, logic and fluency of the generated explanations.

\subsection{Experimental setup}

\begin{figure*}[!t]
    \centering
    \includegraphics[width=\textwidth]{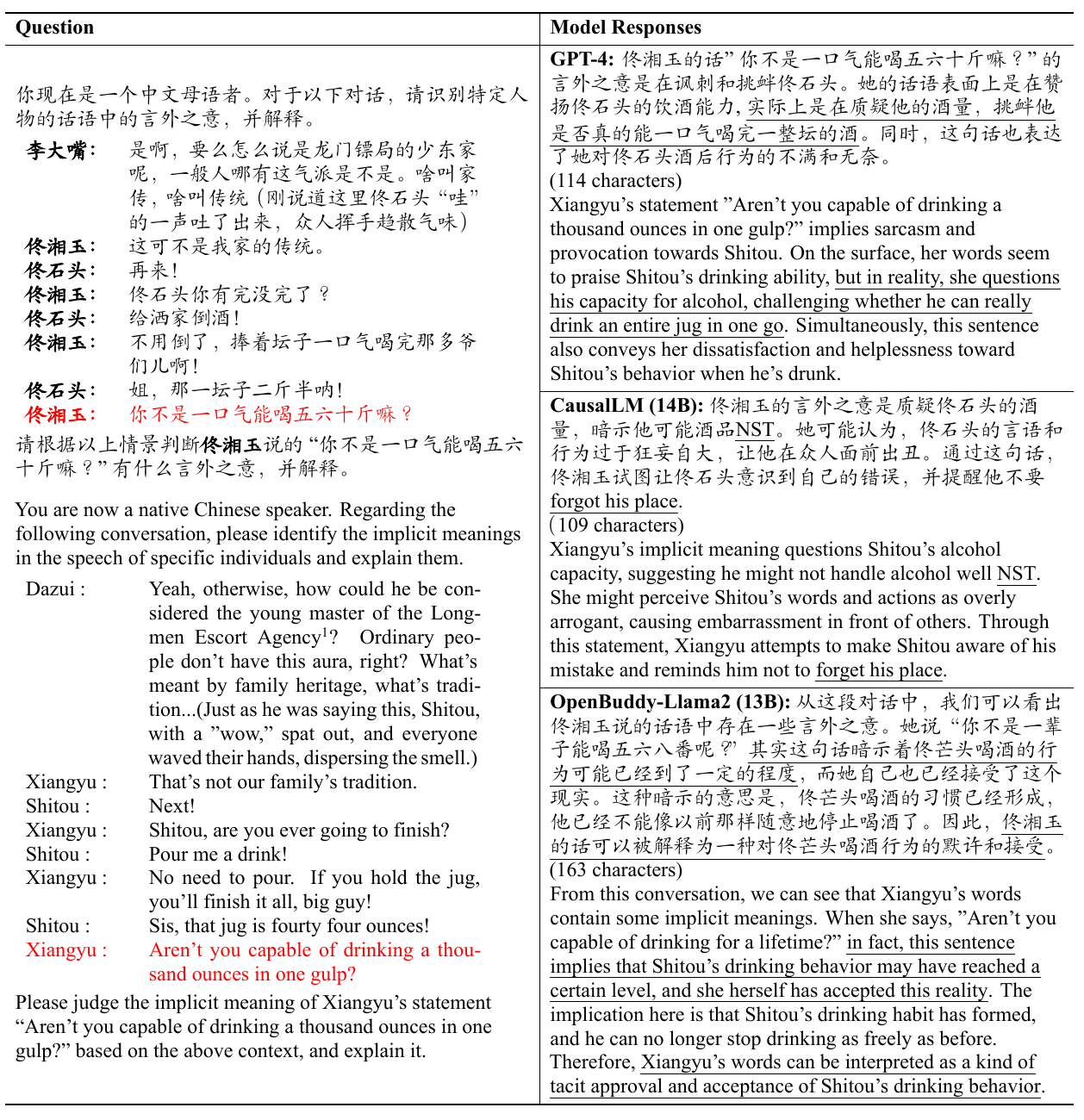}
    \caption{Models' interpretations to a character line (highlighted with red color) that carries extra meanings in a dialogue with multiple turns. Explanations aberrant from the correct interpretation and non-Chinese characters are highlighted with yellow bottom color. Escort Agency refers to historical security firms that were typically hired by trade merchants for protection during transportation of goods. They were known to use martial arts for defense, similar to bodyguards or security personnel.}
    \label{tab:exp2-response-eg}
\end{figure*}

We first performed a pilot trial comprising five questions. Among the eight models, BLOOMZ (7.1B), text-davinci-002 and text-davinci-003 produce short and fragmented responses, despite our prompts explicitly asking for detailed explanation. We therefore select GPT-3.5-Turbo, GPT-4, Chinese-Alpaca-2 (13B), OpenBuddy-Llama2 (13B) and CausalLM (14B) for this experiment. 
We increase \texttt{max\_new\_tokens}
for open-sourced models, and keep the same hyperparameter for GPT models (See Appendix~\ref{sec:appendix:B}). We use the prompt 
in Figure~\ref{tab:exp2-response-eg} to instruct the models to formulate an explanation for the target sentence. This prompt is selected after a series of trial experiments to ensure the quality of models' responses.

\subsection{Human evaluation}
We invite 15 human participants, all of whom are linguistics majors at a top Chinese university, to evaluate the answers generated by models in three dimensions: reasonability or reasonableness, logic and fluency. Reasonability is gauged by the degree to which the response is aligned with the interpretation provided by the human annotator. Logic is assessed based on the presence of a clear and consistent chain of reasoning in the response. Fluency, on the other hand, evaluates whether the generated text is devoid of incoherent elements such as foreign language insertions or unintelligible code fragments.

\subsection{Results}
\label{sec:exp2:res}
Table~\ref{tab:models-scores-in-3D} shows the mean ratings of the responses of the five models' in three dimensions.
Responses from GPT-4 scores the highest in all three dimensions with the smallest variance. Responses from GPT-3.5-Turbo are rated high but with larger standard deviations, suggesting unstable performance.
Answers from the other three models are comparable to each other ($F=.964$, $p=.549$). 
Notably, the scores of CausalLM (14B) are lower than those of GPT-3.5-Turbo, which is inconsistent to results of Experiment 1. 
As providing coherent explanations of the implicatures is much harder than picking one answer from four options, this suggests that a model good at the latter may not necessarily be equally good at the former. 

Examining the generated answers in detail explains the distribution of human rating scores. Text generated by GPT-4 and GPT-3.5-Turbo are both identified with a minimal number of ``code switching", the phenomenon of adulterating non-Chinese tokens into their responses, with the interpretation from GPT-4 being more precise and coherent than that from GPT-3.5-Turbo. While Chinese-Alpaca-2 (13B), OpenBuddy-Llama2 (13B) and CausalLM (14B) have comparable performance, they all underperform the two GPT models. They reflect different patterns of generated text. Both Chinese-alpaca-2 (13B) and Openbuddy-Llama2 (13B) feature highly pure Chinese responses with their scores in text quality above 4.5, but the latter model scores distinctively low in reasonability. 
CausalLM-14B, on the contrary, tends to code-switch or generate
tokens in English frequently.

Figure~\ref{tab:exp2-response-eg} presents a typical example to illustrate the different styles of generation from the models.
The target sentence in this dialogue doesn't reflect a normal judgment based on his alcoholic words and behaviors but rather proposes a totally unreal scenario. The overt absurdity in the words represents an obvious signal of the violation of the maxim of Quality. An implicature from Xiangyu that Shitou can not drink anymore thus arises and is precisely conveyed to the listeners. 
Her words also express her irony and dissatisfaction towards Shitou. 
GPT-4 gives a concise interpretation that aligns closest to the reference interpretation among the selected models. However, it has mistaken the ironic tone as questioning Shitou's capacity for liquor, as highlighted in the example. 
CausalLM (14B) 
produces a correct interpretation by and large, but the quality of its answer is negatively impacted by its poor fluency, in that some English words and meaningless character sequences appear in its answer.
It is interesting that the ``forgot his place" carries the correct meaning, which leads us to consider this as code-switch, rather than nonsensical generation. This code-switching phenomenon appears to be a feature specific to the responses from CausalLM (14B).
The response from Openbuddy-Llama2 (13B) exemplifies 
a response that is both verbose and irrelevant. 

\section{Discussion and Future Work}
\label{sec:disc}

\subsection{LLM's understanding of conversational implicature in Chinese?}

Our results from Experiment 1 show that the performance of GPT4 on our proposed benchmark is on par with humans, while other models are at least 15 points behind (including GPT-3.5-turbo). 
This suggests that while in principle pragmatic implicatures can be acquired by arguable the best LLMs at the moment, it is a non-trivial task for other LLMs.

Results from Experiment 1 also reveal no significant by-maxim variance in human accuracy, as well as model accuracy (see Figure~\ref{fig:by_maxim}). This is different from the results in previous work on human processing of implicatures~\citep{engelhardt2006speakers, rubio2019overinformative, okanda2015understanding, panzeri2021children}, which demonstrate that humans sanction infringements of the maxims in different ways, being less sensitive to the violation of the maxim of quantity than to others, leading to more processing difficulty for this maxim. This difference is possibly because many dialogues in our dataset violate several maxims, rather than a single one.
Thus we do not observe a difference in human processing of implicature derived from violation of different maxims.

Similar in evaluation paradigm, \citet{finegrainedPragmatic} classifies their pragmatic-related questions according to Theory of Mind (ToM), in which the violation of Gricean maxims is a single category without sub-categories. Text-davinci-002 is evaluated in both their and our studies, resulting in different performances. It surpasses 60\% accuracy on the English dataset in \citet{finegrainedPragmatic}, but does not reach 25\% accuracy on our dataset. We attribute this to the difference in our selection of text sources. While \citet{finegrainedPragmatic} uses hand-crafted scenarios targeting humor, sarcasm, and other pragmatic phenomena separately, often with fewer turn-taking, our text source comes from the script of the sitcom, with many more turn-taking, each of which contains multiple phenomena. The degradation in text-davinci-002's performance can serve as a calibration of the difficulty of questions in the two datasets. With text-davinci-002 being the only model with accuracy lower than 40\% in the current study, the results demonstrate a perceptible development in the pragmatic understanding of the newly developed LLMs.

\subsection{Future directions}
Results from Experiment 2 indicates the possibility that a model with high accuracy in multiple-choice question could fail in a free-text generation task of interpreting the pragmatic meaning on its own (CausalLM-14B). 
We thus argue that multiple-choice questions alone is not enough for a comprehensive evaluation of LLMs' linguistic ability. 
Manual inspection of free-form generation is a must for a more robust analysis of model performance. 
We also see the potential of a more sophisticated design to better quantify their free-form explanations of conversational implicature. 

Future work can also create a large-scale dataset composed of conversational implicature embedded in \textit{naturally occurring} dialogues. 
This will require detailed annotation of spoken corpora, which we believe could benefit from the procedure and implicature-selection criteria used in creating our benchmark. 

\section{Conclusion}

In this paper, we present SwordsmanImp, the first fine-grained Chinese dataset to evaluate LLMs' understanding of conversational implicature. 
In two experiments, we evaluate the state-of-the-art language models' pragmatic skills with two tasks. In Experiment 1, we reveal that GPT-4 attains a human-level accuracy in answering multiple-choice questions, with other models lagging behind. We also find that no significant difference exists for both LLMs and human's accuracy with respect to conversational implicatures that violate different maxims. Results from Experiment 2 reflect the different patterns of generated text across LLMs and after human annotation of the quality of generated text from three dimensions, we reveal that while most models produce fluent text, they struggle to generate coherent and sensible explanations for the implicature, even if the model has achieved high accuracy in multiple-choice question. 

\section*{Acknowledgment}
We thank Xinjia Qi, Qiyu Sun and Yaqian Zhang for verifying the implicatures and improving the dataset. We thank all  participants for their support in this study. We also thank the anonymous reviewers for their valuable comments. This project is funded by Shanghai Pujiang Program (22PJC063) awarded to Hai Hu. 

\section*{Limitations}
Our dataset is sourced exclusively from the Chinese sitcom \textit{My Own Swordsman}. Although we performed a rigorous proofreading on our data, selecting dialogues whose interpretation only depends on the local information and providing background knowledge when necessary, there may still be features specific to this sitcom such as the personality of the characters that play a role in determining the implicature, which may influence the generalizability of our conclusion.

\bibliographystyle{conference}
\bibliography{custom}

\appendix
\section{Evaluation paradigms and differences}
\begin{figure*}[t]
 \begin{adjustbox}{max width=0.8\textwidth}
    \centering
    \includegraphics[width=\textwidth]{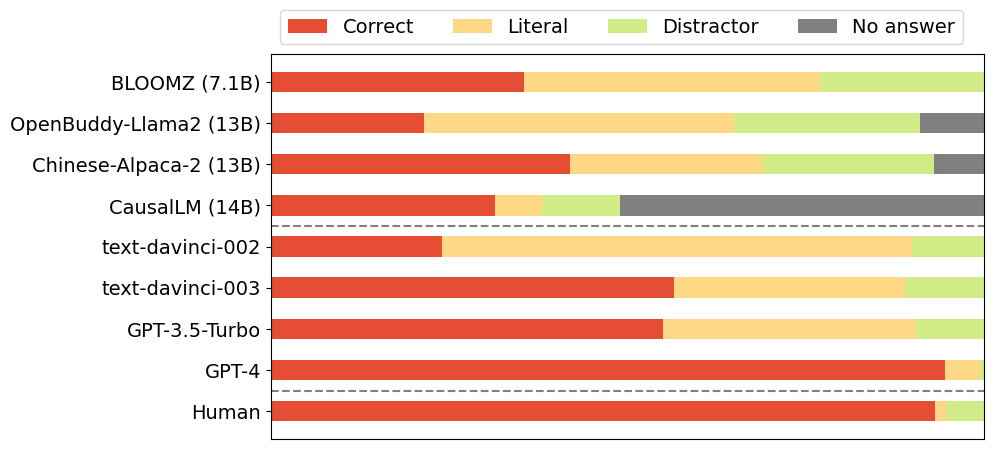}
\end{adjustbox}
    \caption{Answer distribution of models when answers are collected with free generation paradigm.}
    \label{fig:by_choice_old}
\end{figure*}

Models' performance might vary from how their answers are estimated and collected. Next token prediction and free generation are two paradigms separately used to estimate the performance of open-source and close-source models in Experiment 1. Table~\ref{tab:different_strategy} shows the comparison of the open-source models' performance on the multiple-choice questions when their answers are estimated through the two paradigms. The result shows a decrease in accuracy in BLOOMZ (7.1B), CausalLM (13B), and OpenBuddy-Llama2 (13B) and a slight increase in Chinese-Alpaca-2 (13B) when the paradigm switches from next token prediction to free generation. This is aligned with the findings of \citet{li2023cmmlu}. Among the four models CausalLM (13B) has a dramatic decrease in its accuracy, from 78.5\% to 31.5\%, which corresponds to its bad performance in Experiment 2. We find that it fails to give a definite answer in its responses for over half of the questions, as shown in Figure~\ref{fig:by_choice_old}.  

\begin{table*}[t]
    \centering
    \begin{tabular}{ccccccccccc}
    \toprule
        \multirow{3}{*}{Model} & \multicolumn{10}{c}{Accuracy (\%)} \\\cline{2-11}
         & \multicolumn{2}{c}{Total} & \multicolumn{2}{c}{Quantity} & \multicolumn{2}{c}{Quality} & \multicolumn{2}{c}{Relevance} & \multicolumn{2}{c}{Manner}\\
         & Next & Gen & Next & Gen & Next & Gen & Next & Gen & Next & Gen \\
         \midrule
         BlOOMZ & 52.50 &      35.50 &         63.64 &         24.24 &        50.00 &        31.58 &          54.93 &          30.99 &       53.23 &       33.87 \\
        OpenBuddy-Llama2 & 42.50 &      21.50 &         36.36 &         21.21 &        43.42 &        25.00 &          45.07 &          19.72 &       46.77 &       17.74 \\
        Chinese-Alpaca-2 & 41.50 &      42.00 &         42.42 &         36.36 &        39.47 &        44.74 &          45.07 &          43.66 &       45.16 &       38.71 \\
         CausalLM & 78.50 &      31.50 &         81.82 &         42.42 &        80.26 &        31.58 &          76.06 &          30.99 &       88.71 &       40.32 \\
         
    \midrule
         
    \end{tabular}
    \caption{Comparison between the accuracy of open-source models on multiple-choice questions when evaluated with next token prediction and free generation paradigms.}
    \label{tab:different_strategy}
\end{table*}

\section{Hyperparameter setting}
\label{sec:appendix:B}
The hyperparameters used for gathering responses from open-source models in Experiment 1 and Experiment 2 are shown in Table~\ref{tab:hyper1} and Table~\ref{tab:hyper2}.
\begin{table}[h]
\centering
    \begin{tblr}{Q[l, m]Q[l, m]}
    \toprule
        Parameter     &    Value\\
        \midrule
        max\_new\_tokens & 50 \\
        temperature   &    0.9\\
        top\_k        &    3\\
        top\_p        &    0.1\\
        repetition\_penalty & 1.0 \\
        num\_return\_sequence & 1 \\
        do\_sample  & True\\
        \bottomrule
    \end{tblr}
    \caption{Parameter setting for open-source models in Experiment 1}
    \label{tab:hyper1}
\end{table}
\begin{table}[h]
\centering
    \begin{tblr}{Q[l, m]Q[l, m]}
    \toprule
        Parameter     &    Value\\
        \midrule
        max\_new\_tokens & 300 \\
        temperature   &    0.9\\
        top\_k        &    0\\
        top\_p        &    0.9\\
        repetition\_penalty & 1.0 \\
        num\_return\_sequence & 1 \\
        do\_sample  & True\\
        \bottomrule
    \end{tblr}
    \caption{Parameter setting for open-source models in Experiment 2}
    \label{tab:hyper2}
\end{table}

\newpage
\section{Average answer length for different question types}
We present the average answer length for the four maxims in Table~\ref{tab:avg:length:exp2}.

\begin{table*}[t]
\centering
\begin{adjustbox}{max width=0.8\textwidth}
\begin{tabular}{lcccc}
\toprule
&\textbf{Pragmatic}&\textbf{Literal}&\textbf{Distractor\#1}&\textbf{Distractor\#2}\\
\midrule
Total&19.20&16.38&12.54&12.21\\
Quality&18.32&15.88&11.96&11.72\\
Quantity&18.36&16.73&12.30&12.00\\
Relevance&20.77&16.62&13.11&12.77\\
Manner&18.10&17.18&12.13&12.06\\
\bottomrule
\end{tabular}
\end{adjustbox}
\caption{Average number of Chinese characters in each answer for all questions and for each type of questions. \label{tab:avg:length:exp2} }
\end{table*}

\section{Prompt for experiment 1}\label{app:prompt:exp1}

\begin{quote}
\small
    \begin{CJK*}{UTF8}{gkai}你现在是一个中文母语者。对于以下对话，请识别特定人物的话语中的的言外之意，在给出的四个选项中选择一个你认为的正确答案。\end{CJK*}
    [En: You are now a native Chinese speaker. For the following dialogue, please identify the implied meaning in the specific character's speech, and choose one of the four given options that you think is the correct answer.  ]
    
    < Dialogue >
    
    < Four interpretations as choices > 
    
    (Close-source models:) \begin{CJK*}{UTF8}{gkai}请在`Response:'后写出你选择的答案。\end{CJK*} [En: Please write your answer after `Response:']

    (Open-source models:) \begin{CJK*}{UTF8}{gkai}答案：\end{CJK*} [En: Answer:]

\end{quote}

\end{document}